\documentclass[letterpaper]{article} 
\usepackage[draft]{aaai2026}  
\usepackage{times}  
\usepackage{helvet}  
\usepackage{courier}  
\usepackage[hyphens]{url}  
\usepackage{graphicx} 
\usepackage{natbib}  
\usepackage{caption} 
\usepackage{algorithm}
\usepackage{algorithmic}
\usepackage{amsmath}
\usepackage{amssymb}
\usepackage{booktabs} 
\usepackage{multirow}
\usepackage{subcaption}
\usepackage[table]{xcolor} 
\usepackage{microtype}      
\usepackage{enumitem}       

\newcommand{\method}{DaGRPO} 

\DeclareMathOperator{\mean}{mean}
\DeclareMathOperator{\std}{std}
\DeclareMathOperator{\clip}{clip}

\title{DaGRPO: Rectifying Gradient Conflict in Reasoning via \\ Distinctiveness-Aware Group Relative Policy Optimization}

\author{
    Xuan Xie,\textsuperscript{\rm 1}
    Xuan Wang,\textsuperscript{\rm 2}\thanks{Corresponding author:  wangxuan39@meituan.com}
    Wenjie Wang,\textsuperscript{\rm 2}
    Shuai Chen,\textsuperscript{\rm 2}
    Wei Lin\textsuperscript{\rm 2}
}
\affiliations{
    \textsuperscript{\rm 1}Tsinghua University\\
    \textsuperscript{\rm 2}Meituan\\
}

\begin{document}

\maketitle

\begin{abstract}
The evolution of Large Language Models (LLMs) has catalyzed a paradigm shift from superficial instruction following to rigorous long-horizon reasoning. While Group Relative Policy Optimization (GRPO) has emerged as a pivotal mechanism for eliciting such post-training reasoning capabilities due to its exceptional performance, it remains plagued by significant training instability and poor sample efficiency. We theoretically identify the root cause of these issues as the lack of distinctiveness within on-policy rollouts: for routine queries, highly homogeneous samples induce destructive gradient conflicts; whereas for hard queries, the scarcity of valid positive samples results in ineffective optimization. To bridge this gap, we propose \textbf{D}istinctiveness-\textbf{a}ware \textbf{G}roup \textbf{R}elative \textbf{P}olicy \textbf{O}ptimization (\textbf{\method}). \method{} incorporates two core mechanisms: (1) \textit{Sequence-level Gradient Rectification}, which utilizes fine-grained scoring to dynamically mask sample pairs with low distinctiveness, thereby eradicating gradient conflicts at the source; and (2) \textit{Off-policy Data Augmentation}, which introduces high-quality anchors to recover training signals for challenging tasks. Extensive experiments across 9 mathematical reasoning and out-of-distribution (OOD) generalization benchmarks demonstrate that \method{} significantly surpasses existing SFT, GRPO, and hybrid baselines, achieving new state-of-the-art performance (e.g., a +4.7\% average accuracy gain on math benchmarks). Furthermore, in-depth analysis confirms that \method{} effectively mitigates gradient explosion and accelerates the emergence of long-chain reasoning capabilities.
\end{abstract}

\section{Introduction}
The quest for Artificial General Intelligence (AGI) demands that Large Language Models (LLMs) transcend simple pattern matching to master complex reasoning. Unlike standard generation tasks, mathematical and logical reasoning requires models to navigate vast solution spaces and maintain rigorous logical consistency over long horizons—capabilities often referred to as ``System 2'' thinking \citep{li2025system}. Consequently, the post-training paradigm has shifted toward the elicitation of these complex Chain-of-Thought (CoT) capabilities \citep{jaech2024openai,guo2025deepseek,team2025kimi}. While Supervised Fine-Tuning (SFT) provides a foundation, it often fails to generalize beyond the training distribution or correct self-generated errors. This necessitates the adoption of Reinforcement Learning (RL) with Verifiable Rewards to facilitate robust self-improvement. In this landscape, Group Relative Policy Optimization (GRPO) \citep{shao2024deepseekmath} has anchored itself as a pivotal technique. By introducing Group Relative Advantage, GRPO obviates the dependency on a computationally expensive Value Function, significantly reducing memory footprint while effectively driving performance breakthroughs in tasks requiring multi-step logical deduction, such as mathematics and coding.

\begin{figure}[t]
    \centering
    \includegraphics[width=\linewidth]{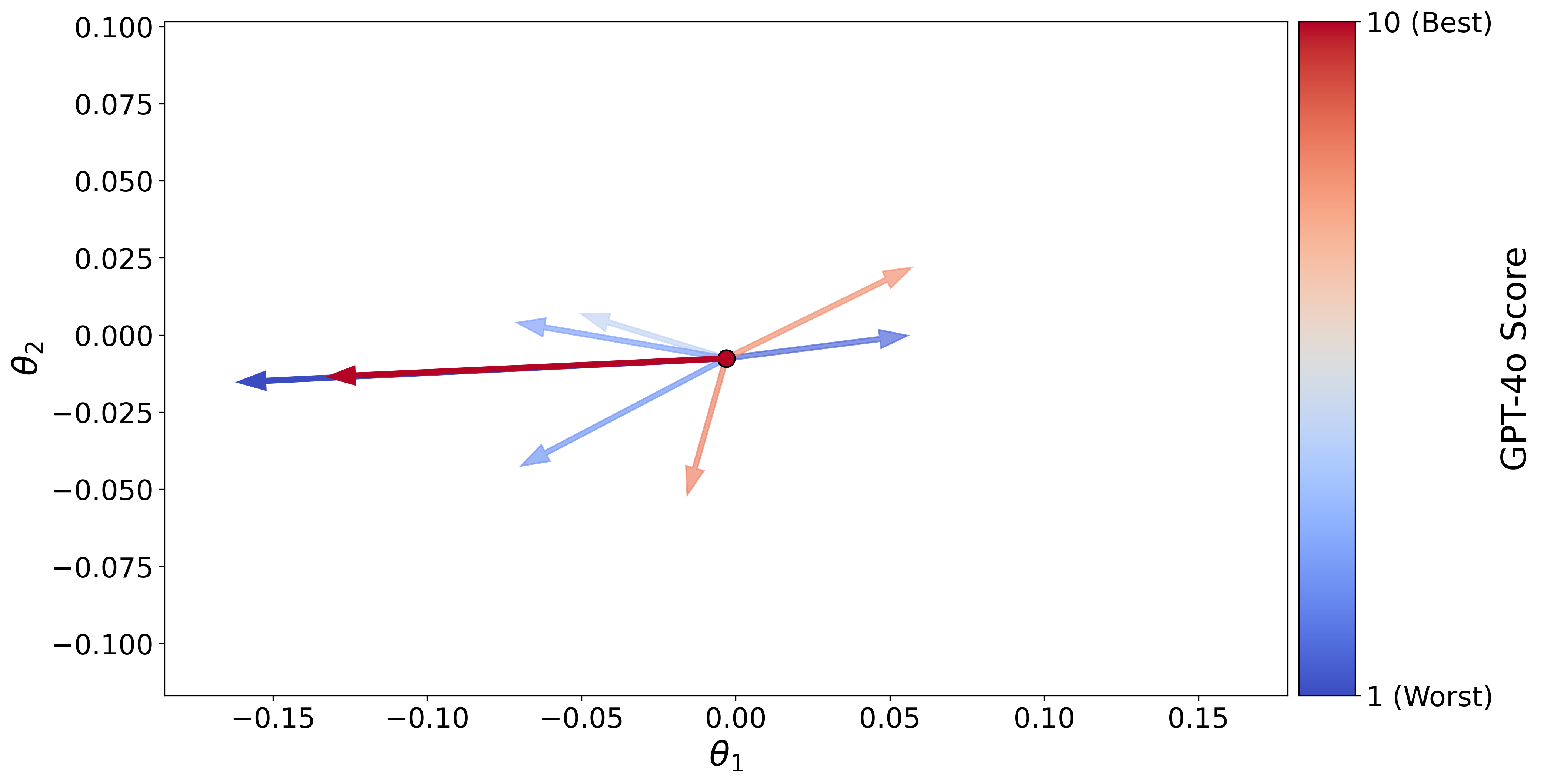}
    \caption{\textbf{Visualization of Gradient Conflict.} Using Qwen2.5-0.5B-Instruct with group size $G=8$, we visualize the gradient directions of distinct responses for an identical prompt. The color spectrum (Red=High Score, Blue=Low Score) reflects GPT-4o evaluations.}
    \label{fig:gradient}
\end{figure}

However, as an on-policy algorithm, GRPO inherently suffers from training instability—a challenge where off-policy counterparts like DPO \citep{rafailov2023direct} demonstrate superior robustness. Prior studies \citep{trott2019keeping,simoni2025gtpo} attribute this instability primarily to two factors: 
\begin{enumerate}[leftmargin=*, noitemsep, topsep=2pt]
    \item \textit{Reward Sparsity}: For challenging queries requiring long-horizon reasoning, the vast majority of on-policy rollouts yield low rewards, resulting in zero policy gradients that contribute negligibly to policy updates. 
    \item \textit{Gradient Conflict}: As illustrated in Figure \ref{fig:gradient}, training Qwen2.5-0.5B-Instruct with a group size of $G=8$ reveals that for an identical prompt, distinct responses within the group exhibit divergent or even diametrically opposed gradient directions on specific parameters during backpropagation. This conflict causes valid gradient signals to cancel each other out or deviate from the optimal trajectory, thereby severely compromising both training stability and sample efficiency. 
\end{enumerate}

However, we argue that these phenomena are merely superficial symptoms. The root cause of GRPO's instability lies in the \textit{lack of distinctiveness} among on-policy rollouts—a deficiency inextricably linked to GRPO's inherent optimization mechanism.. Prior literature \citep{wu2025takes} suggests that GRPO implicitly operates under a Contrastive Learning hypothesis: by normalizing intra-group rewards, the algorithm implicitly categorizes responses into positive (positive advantage) and negative (negative advantage) samples, aiming to boost the probability of the former while suppressing the latter. Crucially, the efficacy of such contrastive optimization hinges on the semantic distinctiveness of the sampled trajectories: effective gradient signals emerge only when high-quality positive anchors are contrasted against sufficiently distinct negative references. 

In practice, however, datasets often provide only input prompts and final ground truths, lacking fine-grained supervision for intermediate deductive steps. Consequently, the on-policy sampling process encounters a twofold dilemma: 
\begin{itemize}[leftmargin=*, noitemsep, topsep=2pt]
    \item \textit{``Homogenized Conflict'' in Easy Tasks}: For routine queries, the majority of on-policy reasoning trajectories generated by the model exhibit high homogeneity. This results in insufficient logical distinctiveness between positive and negative samples. Consequently, critical reasoning tokens shared across these samples simultaneously receive contradictory signals—reinforcing and suppressing—triggering severe gradient conflicts. As validated by GPT-4o scoring in Figure \ref{fig:gradient}, these low-distinctiveness pairs act as noise rather than valid feedback for logical refinement.
    \item \textit{``Benchmark Absence'' in Hard Tasks}: For queries exceeding current capabilities, the model typically generates uniformly poor rollouts with rewards approaching zero. The absence of a high-quality positive anchor within the group renders the entire batch ineffective. In summary, homogenization in easy tasks and quality degradation in hard tasks collectively result in a severe lack of distinctiveness in the reasoning space.
\end{itemize}

To address these challenges, we propose \textbf{\method{}}, a novel framework that explicitly enhances sample distinctiveness to rectify gradient update directions. Specifically, \method{} incorporates two core mechanisms: 
1. \textit{Sequence-level Gradient Rectification}: We introduce a fine-grained scoring mechanism (via LLM-as-a-Judge) to quantify intra-group distinctiveness, dynamically masking ``ambiguous'' pairs that fall below a critical threshold. This eradicates gradient conflicts stemming from low-quality contrastive pairs at the source, ensuring that policy updates are driven exclusively by signals with clear discriminative value. 
2. \textit{Off-policy Data Augmentation}: To mitigate benchmark absence in hard tasks, we employ a hybrid sampling strategy that injects off-policy demonstrations from a stronger model (e.g., DeepSeek-R1) into the on-policy group as high-quality positive anchors. This mechanism not only recovers the gradient signals for challenging queries but also provides explicit logical guidance. 

Extensive evaluations across 6 competition-level mathematical reasoning datasets and 3 out-of-distribution (OOD) benchmarks demonstrate the superiority of our approach. \method{} achieves state-of-the-art performance across all evaluated benchmarks. Specifically, on in-distribution math tasks, \method{} attains an average accuracy of 48.8\%—a substantial +4.7\% gain over the standard GRPO baseline—significantly outperforming strong competitors like Oat-Zero \citep{liu2025understanding} and LUFFY \citep{yan2025learning}. Furthermore, \method{} exhibits superior generalization capabilities and training stability on OOD tasks.

\noindent Our main contributions are summarized as follows:
\begin{itemize}[leftmargin=*, noitemsep]
    \item We provide a comprehensive analysis of gradient conflict and sample inefficiency in GRPO through the lens of contrastive learning. We identify the ``\textit{Lack of Distinctiveness}'' within on-policy rollouts as the fundamental pathology driving optimization instability.
    \item We propose \method{}, a framework that synergizes two core mechanisms: \textit{Sequence-level Gradient Rectification} to dynamically filter low-distinctiveness samples for conflict elimination, and \textit{Off-policy Data Augmentation} to introduce high-quality anchors, thereby recovering training signals for challenging tasks.
    \item We conduct extensive experiments across 9 benchmarks. Empirical results confirm that \method{} effectively mitigates gradient conflicts while significantly pushing the performance boundaries and OOD generalization capabilities of LLMs in complex reasoning tasks.
\end{itemize}

\section{Related Work}

\noindent\textbf{Reinforcement Learning for LLM Post-training.} Reinforcement Learning (RL) has established itself as the cornerstone paradigm for post-training Large Language Models (LLMs), evolving from preference alignment to complex reasoning elicitation. 
Foundational work by \citet{christiano2017deep} and \citet{ouyang2022training} (InstructGPT) solidified PPO \citep{schulman2017proximal} as the standard for aligning models with human intent. While PPO remains effective, its dependency on Value Functions and Reward Models incurs substantial memory overhead, a bottleneck particularly acute in reasoning tasks that require extensive exploration.
To mitigate these costs, Direct Preference Optimization (DPO) \citep{rafailov2023direct} emerged as a reward-free, off-policy alternative, optimizing directly on static preference data. Subsequent variants like IPO \citep{azar2024general} and KTO \citep{ethayarajh2024kto} further refined this approach to address overfitting and data constraints. 
On-policy and off-policy methods present inherent trade-offs: On-policy methods (e.g., PPO) ensure consistency between sampling and policy distributions, fostering better exploration, yet suffer from poor sample efficiency. Conversely, off-policy methods like DPO optimize data utilization but remain susceptible to distribution shift.

\noindent\textbf{Optimization and Analysis of GRPO.} As the core algorithm of DeepSeek-R1 \citep{guo2025deepseek}, Group Relative Policy Optimization (GRPO) replaces the traditional value function with group-wise relative advantage estimation, demonstrating a superior performance-compute ratio in mathematical and code reasoning tasks. However, its widespread adoption has exposed inherent limitations, catalyzing a surge of targeted research. Addressing training instability, DAPO \citep{yu2025dapo} identifies the risk of entropy collapse and proposes clip-higher and dynamic sampling strategies, GSPO \citep{zheng2025group} improves the stability of MoE model training by elevating the optimization granularity from the token level to the sequence level. GTPO \citep{simoni2025gtpo} attempts to mitigate gradient conflict by selectively masking the gradient contributions of conflicting tokens. However, this strategy typically only identifies continuous conflicting tokens at the sequence boundaries (i.e., beginning and end), effectively ignoring the conflicts triggered by insufficient logical distinctiveness deep within the reasoning chains. Concurrently, \citet{wu2025takes} reformulate the GRPO objective from a contrastive learning perspective, revealing its intrinsic mathematical connection with DPO. To further tackle the sample inefficiency of exclusive on-policy sampling, recent works have explored hybrid paradigms incorporating off-policy data. For instance, CHORD \citep{zhang2025policy} utilizes a global coefficient to dynamically balance the gradient weights between SFT data (off-policy) and GRPO samples (on-policy), while LUFFY \citep{yan2025learning} integrates offline trajectories directly into online groups. Different from these approaches, we identify the lack of sample distinctiveness as the root cause of optimization instability and propose a distinctiveness-aware mechanism that dynamically filters homogeneous conflicts and introduces high-quality contrastive references.

\section{Preliminaries} 
Unlike PPO, which relies on a learned Value Function for advantage estimation, GRPO employs a group-based relative estimation strategy that significantly reduces computational overhead. Specifically, for each input prompt $q$, GRPO samples a group of $G$ responses  $\left\{o_i \right\}_{i=1}^G$ from the old policy $\pi_{\theta_{\text{old}}}$, where a reward model assigns a scalar reward $R(o_i)$ to each response. The advantage $A_i$ for the $i$-th response is then derived via intra-group normalization:
\begin{equation}
A_i = \frac{R(o_i) - \mean\left( \{ R(o_i) \mid i = 1, \ldots, G \} \right)}{\std\left( \{ R(o_i) \mid i = 1, \ldots, G \} \right)}
\end{equation}
Building on this, GRPO adopts a clipped surrogate objective similar to PPO, incorporating a KL divergence penalty term as a constraint:
\begin{multline}
    \mathcal{J}_{\text{GRPO}}(\theta) = \mathbb{E}_{\substack{q \sim \mathcal{Q} \\ o_i \sim \pi_{\theta_{\text{old}}}}} \Bigg[ \frac{1}{G} \sum_{i=1}^G \frac{1}{|o_i|} \sum_{t=1}^{|o_i|} \Bigg( \\
    \min\left ( r_{i,t}(\theta) A_i, \clip( r_{i,t}(\theta), 1 - \epsilon, 1 + \epsilon) A_i \right) \\
    - \beta D_{\text{KL}}(\pi_\theta \|\pi_{\text{ref}}) \Bigg) \Bigg].
\end{multline}
where, $r_{i,t}(\theta) = \frac{\pi_\theta(o_{i,t} \mid q, o_{i,<t})}{\pi_{\theta_{\text{old}}}(o_{i,t} \mid q, o_{i,<t})}$. Assuming the clipping operation is inactive and omitting the KL divergence penalty, the GRPO optimization objective reduces to:
\begin{equation}
\mathcal{J}_{\text{GRPO}}(\theta) = \underset{\substack{q \sim \mathcal{Q} \\ o_i \sim \pi_{\theta_{\text{old}}}}}{\mathbb{E}}\left[ \frac{1}{G} \sum_{i=1}^G \frac{1}{|o_i|} \sum_{t=1}^{|o_i|}  r_{i,t}(\theta) \cdot A_i \right],
\end{equation}
Consequently, the simplified policy gradient is derived as:
\begin{equation}
\nabla_\theta \tilde{\mathcal{J}}_{\text{GRPO}} = \frac{1}{G} \sum_{i=1}^G \frac{A_i}{|o_i|} \sum_{t=1}^{|o_i|} r_{i,t}(\theta) \cdot \nabla_\theta  \log\pi_\theta(o_{i,t}|q, o_{i,<t})
\end{equation}
Notably, the advantage $A_i$ is computed at the sequence-level (based on the final reward of the complete response $o_i$), yet during gradient updates, it is uniformly broadcasted as a scalar weight to every token within the sequence. This implies that every token receives an identical reinforcement signal, regardless of its actual contribution to the final reasoning outcome. As discussed subsequently, when sample distinctiveness is low, this ``Granularity Mismatch'' induces severe gradient conflicts, thereby compromising training stability. Since $\pi_{\theta}$ is initialized from $\pi_{\theta_{\text{old}}}$, the probability ratio $r_{i,t}(\theta) \approx 1$ during the initial phase of updates. To facilitate the theoretical analysis of gradient directions in subsequent sections, we adopt the approximation $r_{i,t}(\theta) \approx 1$, focusing primarily on directional conflicts arising from sign differences.

\section{Analysis of Gradient Conflict in GRPO}

Building upon the ``granularity mismatch'' identified in Section 3, we provide a rigorous analysis of how this mismatch precipitates the gradient conflict problem. The core issue stems from the fact that GRPO assigns sequence-level scalar rewards to fine-grained token-level updates, creating ambiguity in credit assignment.

Consider a group of $G$ responses $\{o_i\}_{i=1}^G$ generated for a specific prompt $q$. Based on the sign of the estimated advantage $A_i$, we partition the responses into a positive set $G^+=\{o_i \mid A_i > 0\}$ and a negative set $G^-=\{o_i \mid A_i < 0\}$. The gradient of the GRPO objective can be formally decomposed as:
\begin{equation}
\label{eq:gradient_decomp}
\begin{aligned}
\nabla_\theta \tilde{\mathcal{J}}_{\text{GRPO}} &\approx \underbrace{\frac{1}{G} \sum_{o_j \in G^+} \frac{A_j}{|o_j|} \sum_{t=1}^{|o_j|} g_{j,t}}_{\text{Positive Component } \mathbf{v}^+} + \underbrace{\frac{1}{G} \sum_{o_k \in G^-} \frac{A_k}{|o_k|} \sum_{t=1}^{|o_k|} g_{k,t}}_{\text{Negative Component } \mathbf{v}^-}
\end{aligned}
\end{equation}
where $g_{i,t}=\nabla_\theta  \log\pi_\theta(o_{i,t}|q, o_{i,<t})$ represents the token-level log-probability gradient.

In reasoning tasks, the sampled responses within a group often exhibit significant structural homogeneity, as the policy $\pi_{\theta}$ tends to reuse established reasoning patterns. Let $\mathcal{S}_{\text{shared}}$ denote the set of state-action pairs $(h, w)$—where $w$ is the token and $h$ is the history context—that appear simultaneously in both positive and negative trajectories within the same group. For such a shared instance appearing in $o_j \in G^+$ and $o_k \in G^-$ , the aggregated gradient update for
this specific pattern is:
\begin{equation}
\label{eq:shared_gradient}
\nabla_\theta \tilde{\mathcal{J}}_{\text{shared}} \propto \left( \frac{A_j}{|o_j|} + \frac{A_k}{|o_k|} \right) \nabla_{\theta} \log \pi_{\theta}(w|h)
\end{equation}
Since $A_j > 0$ and $A_k < 0$, these coefficients impose opposing optimization forces on the same parameters. This phenomenon constitutes a form of \textit{destructive interference} \citep{yu2020gradient}, which inevitably leads to severe training instability.

Specifically, when the magnitudes of the advantages are comparable (i.e., $|A_j|/|o_j| \approx |A_k|/|o_k|$), the effective gradient $\nabla_\theta \tilde{\mathcal{J}}_{\text{shared}}$ approaches zero. This neutralizes the learning signal for essential structural knowledge (e.g., reasoning templates or grammatical correctness), forcing the model to ``unlearn'' or ignore valid patterns simply because they co-occurred with incorrect reasoning steps in negative samples. On the other hand, these counteracting updates create a ``tug-of-war'' dynamic, causing the optimization trajectory to oscillate violently in the parameter space (zig-zagging) rather than following a consistent descent direction. This prevents the policy from settling into a stable optimum and significantly hinders convergence efficiency.

\section{Method}
As demonstrated in Section 4, the gradient conflict in GRPO fundamentally arises when shared tokens---representing identical logical steps---simultaneously receive conflicting high-magnitude advantage signals ($A_i > 0$, $A_k < 0$) despite lacking semantic distinctiveness. To mitigate this, we propose a Sequence-level Gradient Rectification mechanism. The core idea is to impose a ``distinctiveness margin'' on the policy update: we explicitly mask out sample pairs that fail to exhibit a sufficient quality gap, ensuring that the policy optimization is driven solely by samples with clear distinguishability.

\noindent \textbf{Fine-grained Quality Assessment.} 
While GRPO relies on binary rule-based rewards, these coarse signals fail to capture the nuance of reasoning quality. For each generated response $o_i$, we employ a LLM-as-a-Judge to compute a fine-grained scalar score $S(o_i) \in [1, 10]$. 
To ensure robust evaluation, we retrieve a high-quality demonstration $o_{\text{demo}}$ (generated by DeepSeek-R1 and verified by math-verify) for the same prompt $q$ as a reference. We prompt GPT-4o to score $o_i$ against $o_{\text{demo}}$ based on three dimensions: correctness of the final answer, validity of the reasoning process, and completeness of the logical steps:
\begin{equation}
S(o_i) = F_{\text{judge}} \left( q, o_i, o_{\text{demo}} \right),\quad i = 1, \dots, G
\end{equation}

\noindent \textbf{Distinctiveness-aware Masking.} 
To filter out ambiguous samples, we establish a dynamic gating mechanism based on the score margin between the positive set $G^+$ and the negative set $G^-$. Let $S_{\text{max}-} = \max\bigl\{ S(o_k) \mid o_k \in G^- \bigr\}$ be the score of the ``best'' negative sample, and $S_{\text{min}+} = \min\bigl\{ S(o_j) \mid o_j \in G^+ \bigr\}$ be the score of the ``worst'' positive sample. We define rectification masks $\lambda^+(\cdot)$ and $\lambda^-(\cdot)$ using a distinctiveness threshold $\delta$:
\begin{enumerate}
    \item \textbf{Positive Mask $\lambda^+(j)$:} A positive sample $o_j$ is retained only if its quality strictly exceeds the best negative sample by the margin $\delta$.
    \begin{equation}
        \lambda^+(j) = \mathbb{I}\bigl[ S(o_j) - S_{\text{max}-} \geqslant \delta \bigr]
    \end{equation}
    \item \textbf{Negative Mask $\lambda^-(k)$:} A negative sample $o_k$ is retained only if its quality is strictly lower than the worst positive sample by the margin $\delta$.
    \begin{equation}
        \lambda^-(k) = \mathbb{I}\bigl[ S_{\text{min}+} - S(o_k) \geqslant \delta \bigr]
    \end{equation}
\end{enumerate}
where $\mathbb{I}[\cdot]$ is the indicator function. This mechanism effectively shields the gradient from ``borderline'' samples that cause confusion. 

Incorporating these masks into the GRPO framework, the rectified policy gradient becomes:
\begin{equation}
\begin{split}
    \nabla_\theta \tilde{\mathcal{J}}_{\text{Rect}} = \frac{1}{G} \Bigg[ & \sum_{o_j \in G^+}  \frac{\lambda^+(j)A_j}{|o_j|} \sum_{t=1}^{|o_j|} g_{j,t} \\
    & + \sum_{o_k \in G^-}  \frac{\lambda^-(k)A_k}{|o_k|} \sum_{t=1}^{|o_k|} g_{k,t} \Bigg]
\end{split}
\end{equation}
By dynamically masking low-distinctiveness samples, we significantly alleviate gradient conflicts throughout the optimization process.

\noindent \textbf{Off-policy Data Augmentation.}
For complex reasoning tasks, models often suffer from exploration collapse. Due to capability limitations, pure on-policy sampling may fail to generate valid reasoning paths, resulting in near-zero rewards for the entire group (i.e., $\forall i, R(o_i) \approx 0$). Consequently, relative advantages vanish ($A_i \approx 0$), rendering the gradient signal uninformative. Furthermore, even if the model occasionally hits the correct final answer via spurious correlations, the reasoning process might be flawed. Reinforcing such low-quality ``lucky guesses'' leads to overfitting and performance degradation. 

Therefore, it is imperative to integrate high-quality off-policy demonstrations (generated by an external strong policy, e.g., DeepSeek-R1) to serve as explicit reasoning anchors. These anchors provide a clear logical reference, guiding the optimization when on-policy exploration fails. We employ a hybrid strategy similar to LUFFY \citep{yan2025learning} to construct a mixed sample group:
\begin{equation}
G_{\text{mix}} = G_{\text{on}} \cup G_{\text{off}}
\end{equation}
Within this mixed group, we re-calculate the standardized advantages. Crucially, off-policy samples typically yield higher rewards, naturally serving as high-quality positive anchors to drive policy updates, while failed on-policy rollouts serve as negatives. The advantage is computed over the union:
\begin{equation}
A_i = \frac{R(o_i) - \text{mean}(\{ R(o_i) \mid o_i \in G_{\text{mix}} \})}{\text{std}(\{ R(o_i) \mid o_i \in G_{\text{mix}} \})}
\end{equation}
As training progresses and the model acquires the capability to generate correct responses independently, on-policy samples naturally gain dominance in the positive set, thereby encouraging autonomous exploration.

Combining the Distinctiveness Masking with the Hybrid Update, the final gradient for \method{} aggregates contributions from the entire mixed batch, applying masks dynamically based on the unified score distribution:
\begin{equation}
\label{eq:final_gradient}
\begin{split}
    \nabla_\theta \tilde{\mathcal{J}}_{\text{DaGRPO}} = \frac{1}{|G_{\text{mix}}|} \Bigg[ & \sum_{o_j \in G^+_{\text{mix}}} \frac{\lambda^+(j)A_j}{|o_j|} \sum_{t=1}^{|o_j|} g_{j,t} \\
    & + \sum_{o_k \in G^-_{\text{mix}}} \frac{\lambda^-(k)A_k}{|o_k|} \sum_{t=1}^{|o_k|} g_{k,t} \Bigg]
\end{split}
\end{equation}
Crucially, the distinctiveness-aware masking described in Eq. (\ref{eq:final_gradient}) applies to the \textit{mixed} group $G_{\text{mix}}$. This provides a safety mechanism against noisy or low-quality off-policy data. If an injected anchor $o_{\text{off}}$ is factually incorrect (low score) or only marginally better than the current policy's best response (score gap $< \delta$), the mask $\lambda$ will automatically zero out its contribution. Consequently, \method{} is robust to the quality variance of external demonstrators, ensuring that the policy only learns from anchors that provide a strictly superior and unambiguous reasoning signal.

\section{Experimental Setups}
\subsubsection{Training Datasets}
We utilize OpenR1-Math-46k-8192 \citep{yan2025learning} as our training dataset, a subset of OpenR1-Math-220k \citep{face2025open} comprising approximately 46,000 prompt-response pairs, with reference responses generated by the expert model DeepSeek-R1 \citep{guo2025deepseek}. Before training, the dataset has been filtered to remove responses exceeding 8k tokens and responses verified as incorrect by math-verify. Within the \method{} framework, these demonstration responses serve a dual purpose: acting as reference answers for the LLM-as-a-Judge mechanism and as high-quality reasoning anchors for Off-policy Data Augmentation.

\subsubsection{Benchmarks}
We evaluate our method on a comprehensive suite of 9 benchmarks, categorized into in-distribution mathematical reasoning and out-of-distribution (OOD) generalization tasks:

\begin{itemize}[leftmargin=*, noitemsep]
    \item \textbf{Mathematical Reasoning (6 Datasets):} To rigorously assess mathematical capabilities, we test on AIME 2024 \citep{aime24}, AIME 2025 \citep{aime25}, AMC \citep{li2024numinamath}, MATH--500 \citep{hendrycks2021measuring}, Minerva \citep{lewkowycz2022solving}, and OlympiadBench \citep{he2024olympiadbench}. Following standard practices for datasets with limited sample sizes (AIME and AMC), we report the avg@32 metric. For the remaining larger benchmarks, we report pass@1.

    \item \textbf{OOD Generalization (3 Datasets):} To verify that our method improves general reasoning beyond math without catastrophic forgetting, we evaluate on ARC--c \citep{clark2018think}, GPQA--Diamond \citep{rein2024gpqa} (expert--level biology, physics, and chemistry), and MMLU--Pro \citep{wang2024mmlu} (interdisciplinary complex reasoning across 57+ subjects).

    \item \textbf{Evaluation Settings:} For all evaluations, we set the sampling temperature to 0.6, top--$p$ to 1.0, and the maximum response length to 8,192 tokens.
\end{itemize}

\subsubsection{Baselines}
To demonstrate the superiority of \method{}, we used Qwen2.5-Math-7B \citep{yang2024qwen2} as the base model and compared it with various strong baseline methods. The baselines are classified as follows: \textbf{Supervised Fine-Tuning (SFT):} Standard supervised fine-tuning using prompts and expert responses in the OpenR1-Math-46k dataset. \textbf{On-policy RL methods:} Original GRPO training based on binary rule rewards. \textbf{Hybrid SFT+RL methods}, including: SFT-GRPO: A two-stage method, first performing SFT fine-tuning, and then GRPO reinforcement learning optimization; LUFFY \citep{yan2025learning}: A hybrid policy optimization method that mixes On-policy and Off-policy data, which we reproduced under the same settings. \textbf{Previous RLVR methods:} We also compared previous Reinforcement Learning with Verifiable Rewards (RLVR) methods, including SimpleRL-Zero \citep{zeng2025simplerl}, OpenReasoner-Zero \citep{hu2025open}, PRIME-Zero \citep{cui2025process}, and Oat-Zero \citep{liu2025understanding}.

\subsubsection{Implementation Details}
Following the configuration of Dr.GRPO \citep{liu2025understanding}, we removed the length normalization and standard deviation normalization terms in advantage estimation, and removed the KL divergence constraint to encourage the model to fully explore. For each prompt, we sample $G=8$ rollouts, with sampling temperature set to 1.0, top-p set to 1.0, and maximum length of 8,192 tokens. We use a global rollout batch size of 128 and an update batch size of 64. The learning rate is set to $1 \times 10^{-6}$. All experiments are conducted for 500 training steps. We adopt a strict binary reward mechanism: the reward is 1 if the final answer is correct, otherwise 0. No additional format rewards or length rewards are used. On this basis, we prompt GPT-4o to perform fine-grained scoring for each rollout, and based on experience, we set the distinctiveness threshold $\delta$ to 3.

\section{Experimental Results}

\begin{table*}[t]
\centering
\caption{Main results on in-distribution mathematical reasoning and out-of-distribution generalization benchmarks. We compare DaGRPO with strong baselines including standard SFT, On-policy RL (GRPO), Hybrid methods (SFT-GRPO, LUFFY), and recent RLVR approaches (SimpleRL-Zero, Oat-Zero, etc.). ``DaGRPO (w/o Off-policy)'' denotes our variant equipped solely with the distinctiveness-aware gradient rectification mechanism. The best and second-best results are highlighted in \textbf{bold} and \underline{underlined}, respectively. * means the results are taken from the corresponding paper.}
\label{tab:main_results}
\resizebox{\linewidth}{!}{
\begin{tabular}{lcccccc|cccc}
\toprule
\textbf{Model} & \multicolumn{6}{c}{\textbf{In-Distribution Performance}} & \multicolumn{4}{c}{\textbf{Out-of-Distribution Performance}} \\
\cmidrule(lr){2-7} \cmidrule(lr){8-11}
& \textbf{AIME 24/25} & \textbf{AMC} & \textbf{MATH-500} & \textbf{Minerva} & \textbf{Olympiad} & \textbf{Avg} & \textbf{ARC-c} & \textbf{GPQA} & \textbf{MMLU-Pro} & \textbf{Avg} \\
\midrule
Qwen2.5-Math & 10.4/3.3 & 29.8 & 47.6 & 9.1 & 16.1 & 19.4 & 10.0 & 12.6 & 14.4 & 12.3 \\
Qwen2.5-Math-Instruct & 12.5/10.2 & 48.5 & 80.4 & 32.7 & 41.0 & 37.6 & 70.3 & 24.7 & 34.1 & 43.0 \\
\midrule
SimpleRL-Zero$^*$  & 27.0/6.8 & 54.9 & 76.0 & 25.0 & 34.7 & 37.4 & 30.2 & 23.2 & 34.5 & 29.3 \\
OpenReasoner-Zero$^*$ & 16.5/15.0 & 52.1 & 82.4 & 33.1 & 47.1 & 41.0 & 66.2 & 29.8 & \textbf{58.7} & 51.6 \\
PRIME-Zero$^*$ & 17.0/12.8 & 54.0 & 81.4 & 39.0 & 40.3 & 40.7 & 73.3 & 18.2 & 32.7 & 41.4 \\
Oat-Zero$^*$ & \textbf{33.4}/11.9 & 61.2 & 78.0 & 34.6 & 43.4 & 43.7 & 70.1 & 23.7 & 41.7 & 45.2 \\
\midrule
SFT & 22.2/\textbf{22.3} & 52.8 & 82.6 & \textbf{40.8} & 43.7 & 44.1 & 75.2 & 24.7 & 42.7 & 47.5 \\
GRPO & 22.7/14.8 & 58.5 & 81.8 & \underline{40.0} & 46.7 & 44.1 & \underline{82.2} & 40.4 & 49.2 & 57.2 \\
SFT $\rightarrow$ GRPO & 25.7/\underline{21.6} & 62.2 & 84.6 & 38.2 & 46.8 & 46.5 & 67.7 & 30.8 & 40.5 & 49.3 \\
\midrule
LUFFY & 23.0/20.8 & \textbf{63.4} & \underline{85.4} & 35.6 & \underline{54.3} & \underline{47.1} & 81.2 & 38.8 & 52.8 & 57.6 \\
DaGRPO (w/o Off-policy) & 23.2/16.1 & \underline{63.1} & 85.0 & 36.7 & 51.2 & 45.9 & 81.3 & \textbf{45.9} & 52.4 & \textbf{59.9} \\
DaGRPO & \underline{29.5}/20.9 & 63.0 & \textbf{86.0} & 37.5 & \textbf{55.8} & \textbf{48.8} & \textbf{83.1} & \underline{40.9} & \underline{53.0} & \underline{59.0} \\
\bottomrule
\end{tabular}
}
\end{table*}

\begin{figure*}[t!]
    \centering
    \setlength{\tabcolsep}{2pt} 
    
    \begin{subfigure}[b]{0.32\linewidth}
        \centering
        \begin{tabular}{@{} m{12pt} m{\dimexpr\linewidth-18pt} @{}}
            \rotatebox{90}{\small \textbf{Outcome Rewards}} & 
            \includegraphics[width=\linewidth]{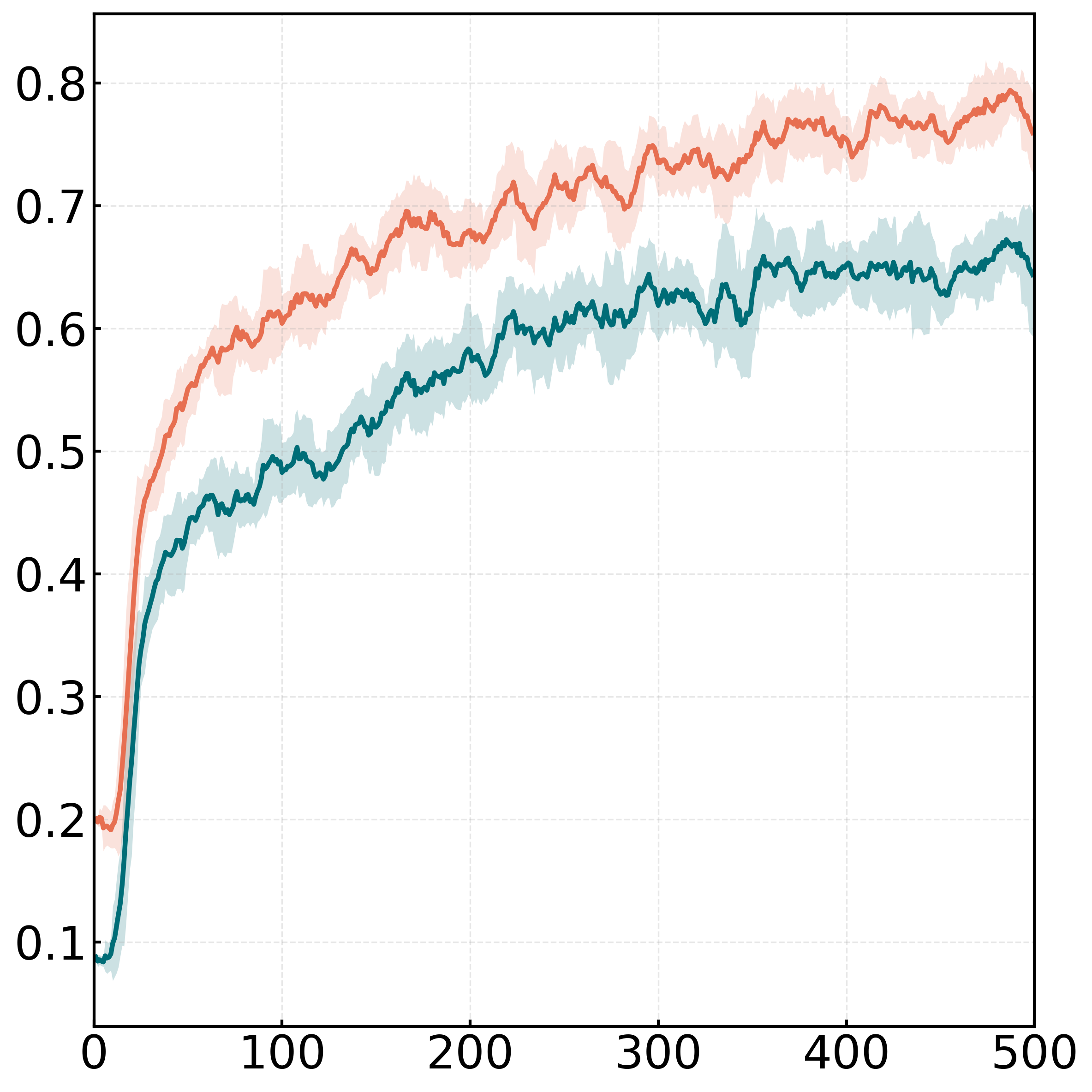} \\
             & \centering \small \textbf{Steps}
        \end{tabular}
    \end{subfigure}
    \hfill
    \begin{subfigure}[b]{0.32\linewidth}
        \centering
        \begin{tabular}{@{} m{12pt} m{\dimexpr\linewidth-18pt} @{}}
            \rotatebox{90}{\small \textbf{Response Length}} & 
            \includegraphics[width=\linewidth]{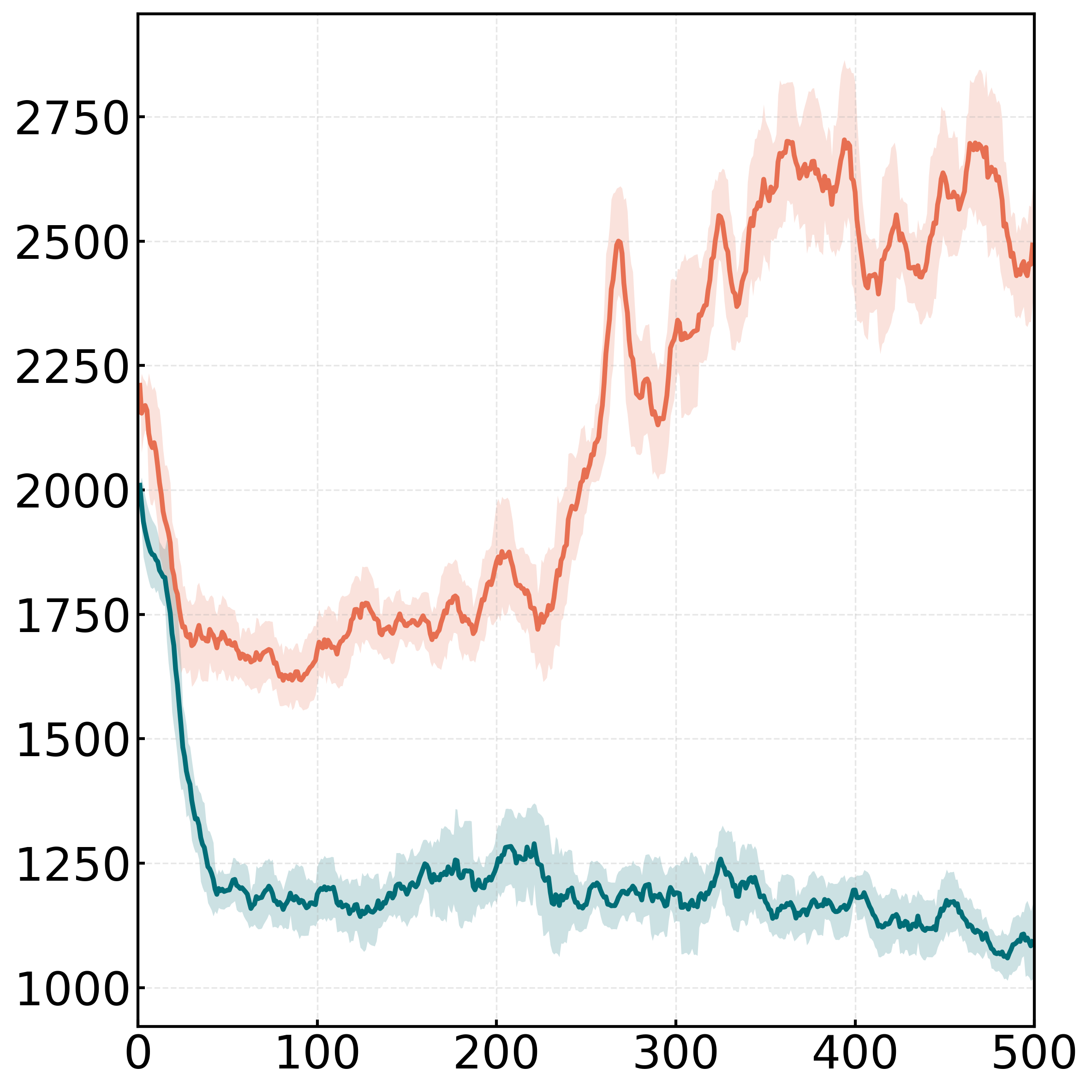} \\
             & \centering \small \textbf{Steps}
        \end{tabular}
    \end{subfigure}
    \hfill
    \begin{subfigure}[b]{0.32\linewidth}
        \centering
        \begin{tabular}{@{} m{12pt} m{\dimexpr\linewidth-18pt} @{}}
            \rotatebox{90}{\small \textbf{Entropy}} & 
            \includegraphics[width=\linewidth]{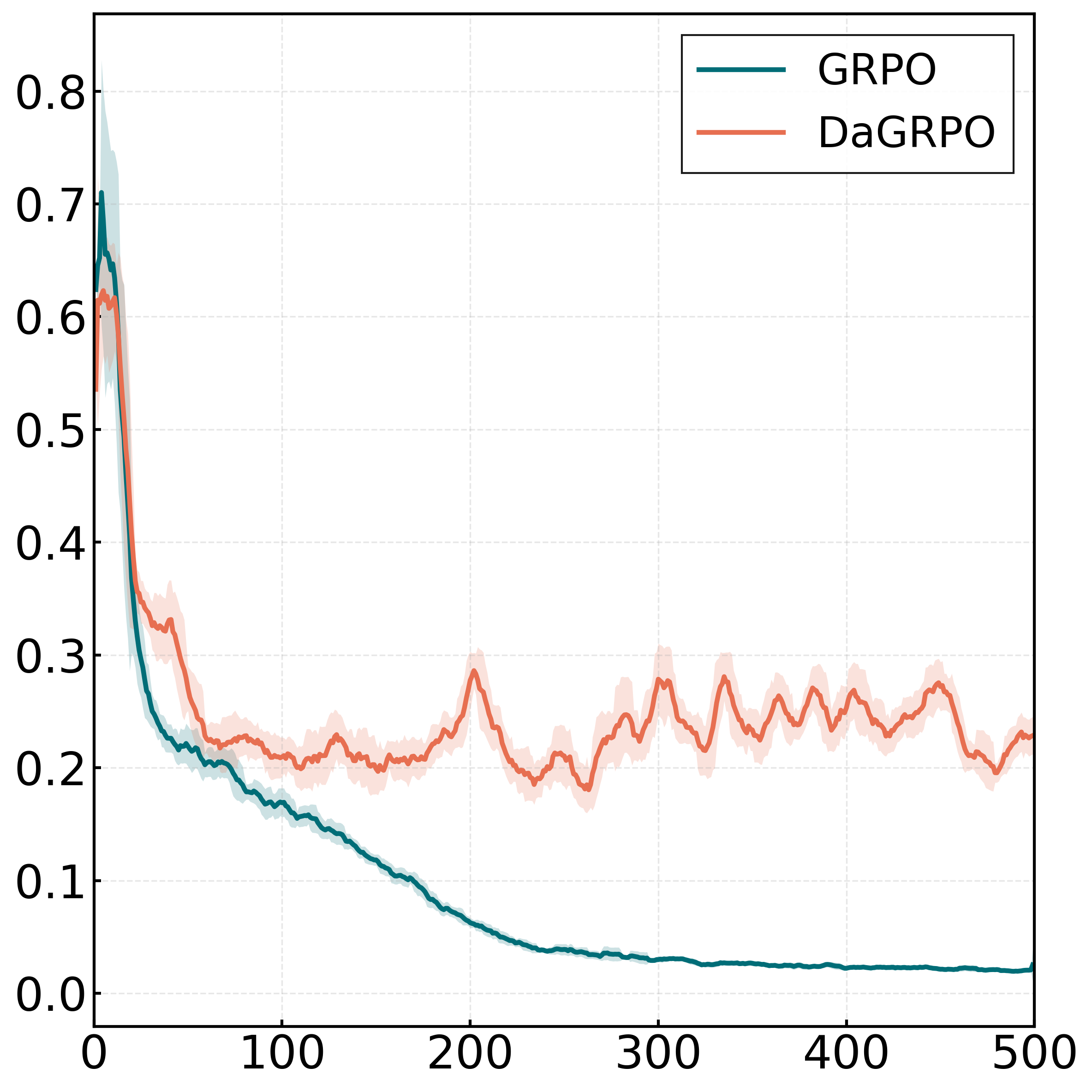} \\
             & \centering \small \textbf{Steps}
        \end{tabular}
    \end{subfigure}

    \caption{\textbf{Training Dynamics.} Comparison of training curves between \method{} and baseline GRPO over 500 steps. \textbf{Left:} Average outcome rewards per step. \textbf{Middle:} Average response length (tokens). \textbf{Right:} Policy entropy.}
    \label{fig:three_metrics}
\end{figure*}

\subsection{Main Results}
\noindent \textbf{Superior Performance on In--Distribution Reasoning.}
As presented in Table \ref{tab:main_results}, \method{} establishes a new state--of--the--art performance among the compared methods. On the aggregate of 6 mathematical reasoning benchmarks, \method{} achieves the highest average accuracy of \textbf{48.8\%}, significantly outperforming the standard GRPO baseline (44.1\%) by a substantial margin of \textbf{+4.7\%}. Furthermore, \method{} consistently surpasses competitive RLVR baselines, such as Oat--Zero (43.7\%), and outperforms the hybrid strategy LUFFY (47.1\%), demonstrating the effectiveness of our distinctiveness--aware optimization framework over simple data mixing strategies.

\noindent \textbf{Efficacy of Gradient Rectification (Mechanism 1).}
The results of the \method{} (w/o Off-policy) variant provide strong empirical support for our gradient rectification mechanism. Even without the injection of off--policy anchors, simply filtering out low--distinctiveness samples yields an average ID score of \textbf{45.9\%}, which already surpasses both the SFT (44.1\%) and GRPO (44.1\%) baselines. This validates our theoretical analysis in Section 4: mitigating gradient conflicts from ambiguous samples is sufficient to stabilize training and unlock performance gains inherent in on--policy exploration.

\noindent \textbf{Critical Role of Anchors in Hard Tasks (Mechanism 2).}
The contribution of the Off--policy Anchor Injection becomes evident in challenging tasks. Comparing the full \method{} with the variant without off-policy data, we observe significant improvements on difficult benchmarks such as \textbf{AIME 2024} (+6.3\% gain from 23.2\% to 29.5\%) and \textbf{OlympiadBench} (+4.6\% gain from 51.2\% to 55.8\%). This confirms that in scenarios where reward sparsity limits on--policy exploration, the high--quality anchors successfully bridge the gap, guiding the model to master complex reasoning paths that are otherwise inaccessible.

\noindent \textbf{Robust Out–of–Distribution Generalization.} 
A key observation is that DaGRPO significantly outperforms baselines on OOD tasks (Avg 59.9\% vs. GRPO 57.2\%). We attribute this to the preservation of policy entropy shown in Figure \ref{fig:three_metrics} (Right). Standard GRPO tends to collapse into narrow, repetitive solution templates (low entropy) to maximize rewards on in-distribution math problems, which harms generalization. By masking out ambiguous samples that encourage ``safe'' but mediocre responses, DaGRPO maintains a higher exploration entropy, preventing overfitting to specific prompt-response patterns and preserving the model's broad reasoning capabilities for unseen domains like biology (GPQA) and science (ARC-c).

\subsection{Training Dynamics Analysis}

To delve deeper into the optimization stability and emergent behaviors of \method{}, we visualize the training dynamics in Figure \ref{fig:three_metrics}. The comparison reveals three critical advantages of our method:

\noindent \textbf{Enhanced Sample Efficiency and Convergence.}
As shown in the outcome reward curve (Figure \ref{fig:three_metrics} Left), \method{} exhibits a significantly faster convergence rate and a higher asymptotic performance compared to GRPO. The initial ``jump--start'' (Step 0--50) can be attributed to the injection of high--quality off--policy anchors, which provide immediate, valid gradient signals. Throughout the training, \method{} maintains a consistent performance gap over GRPO, validating that our gradient rectification mechanism effectively filters out noisy gradients, allowing the model to focus on learning from high--distinctiveness signals.

\noindent \textbf{Emergence of Long--Chain Reasoning.}
The response length evolution (Figure \ref{fig:three_metrics} Middle) presents the most striking difference. While GRPO suffers from ``length collapse''---quickly converging to shorter, potentially shortcut--based responses ($\sim$1200 tokens)--- \method{} exhibits a distinct ``U--shaped'' trajectory. After an initial adjustment, \method{} undergoes a phase transition around Step 200, where the response length surges to over 2500 tokens. Suggesting that \method{} encourages the model to engage in deeper thinking processes. Instead of taking heuristic shortcuts, the model learns to expand its reasoning steps to ensure logical consistency.

\noindent \textbf{Mitigation of Mode Collapse.}
The entropy plot (Figure \ref{fig:three_metrics} Right) explains the mechanism behind the aforementioned behaviors. GRPO's entropy rapidly collapses to near--zero ($<0.05$), indicating severe \textbf{mode collapse} where the policy becomes deterministic and ceases exploration. In contrast, \method{} maintains a healthy level of entropy ($\sim$0.25) throughout training. By masking out ambiguous samples that often drive the policy towards safe but suboptimal modes, \method{} sustains a ``Buffer Zone'' for exploration, enabling the model to continuously discover and refine diverse reasoning paths.

\begin{figure}[t!]
    \centering
    \setlength{\tabcolsep}{2pt} 
    \begin{tabular}{@{} m{12pt} m{\dimexpr\linewidth-18pt} @{}}
        \rotatebox{90}{\small \textbf{Grad Norm}} & 
        \includegraphics[width=\linewidth]{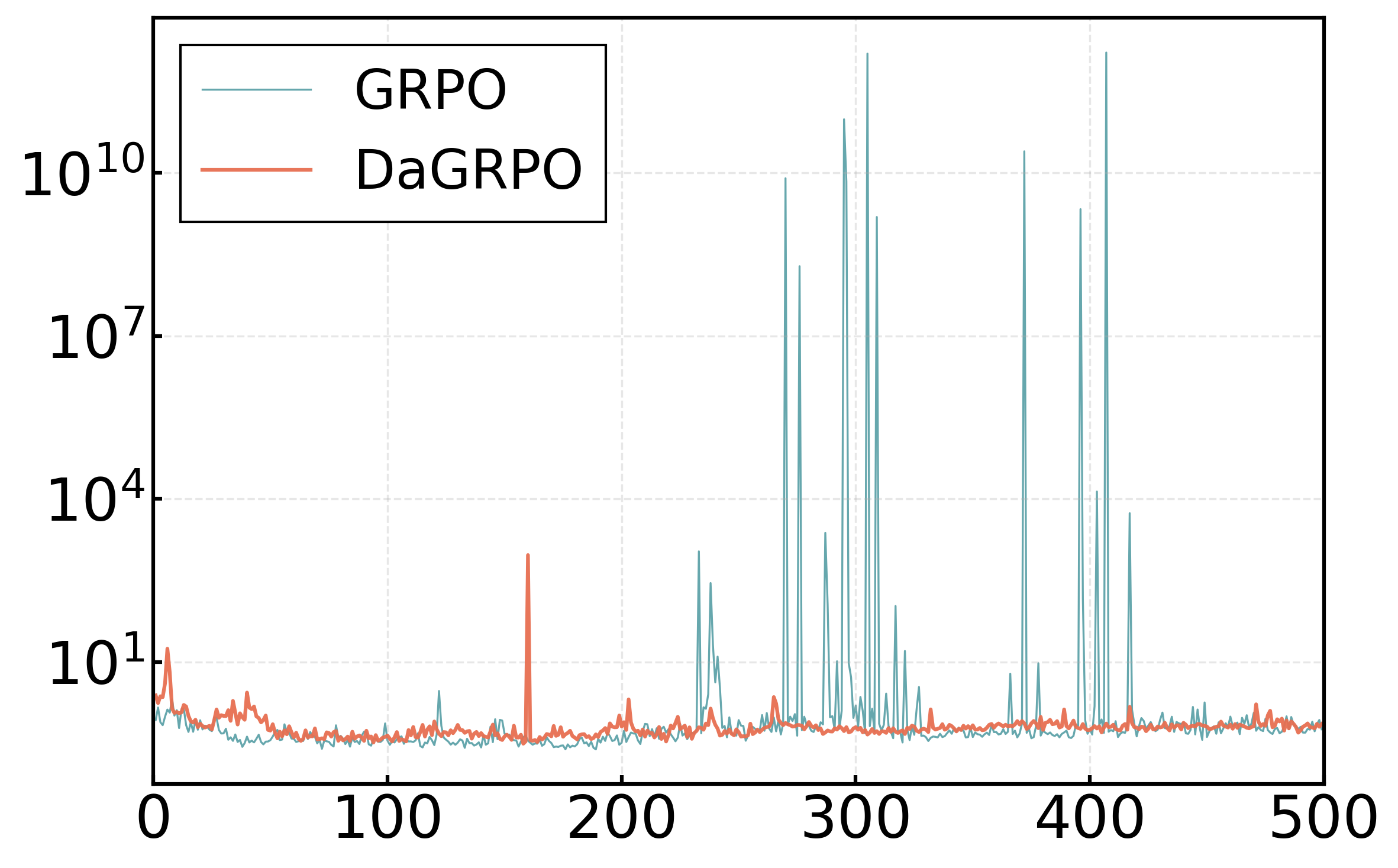} \\
        & \centering \small \textbf{Steps} 
    \end{tabular}
    \caption{Evolution of Gradient Norms throughout the training process. The y-axis is plotted on a logarithmic scale.}
    \label{fig:single_metric}
\end{figure}

\noindent \textbf{Optimization Stability.}
Figure \ref{fig:single_metric} offers visual proof of \method{}'s robustness via gradient norm analysis. Standard GRPO suffers from severe numerical instability, frequently exhibiting gradient explosions of magnitude $10^{10}$, which directly corroborates the destructive impact of gradient conflicts. Conversely, \method{} maintains a smooth and bounded gradient flow by filtering out low-distinctiveness samples. This ``denoising'' mechanism effectively eliminates destructive interference, ensuring a highly stable optimization process.

\section{Conclusion}
In this paper, we addressed the challenges of gradient conflict and low sample efficiency in the post-training of LLM reasoning capabilities by proposing Distinctiveness-aware Group Relative Policy Optimization (\method{}). Through rigorous theoretical analysis, we identified that the root cause of instability in GRPO lies in the lack of semantic distinctiveness within self-sampled reasoning trajectories, which leads to destructive gradient interference. By introducing a sequence-level gradient rectification mechanism to filter low-distinctiveness samples and integrating off-policy data augmentation, \method{} successfully smoothes the optimization trajectory and recalls training signals for hard tasks. Empirical results demonstrate that \method{} achieves state-of-the-art performance across multiple mathematical reasoning and out-of-distribution benchmarks, exhibiting superior training stability and the emergence of long-chain reasoning.

Despite its promising performance, \method{} has certain limitations. First, the gradient rectification mechanism relies on an LLM-as-a-Judge for fine-grained scoring, which introduces additional computational overhead and latency during the training phase. However, this is not a permanent bottleneck; future work can distill the judge's scoring capability into a lightweight, regression-based Reward Model (RM) to replace the online API calls, significantly reducing training latency while maintaining distinctiveness sensitivity. Second, the off-policy augmentation component depends on high-quality external expert data (e.g., from DeepSeek-R1) acting as anchors. In scenarios where strong prior models are unavailable, generating high-distinctiveness sample pairs solely through self-bootstrapping remains a challenging open problem. We leave these explorations for future work to further advance efficient and autonomous reasoning alignment.

\bibliography{references}

\end{document}